\documentclass{article}

\usepackage{arxiv}

\usepackage[utf8]{inputenc}
\usepackage{hyperref}       
\usepackage{url}            
\usepackage{amsfonts}
\usepackage{algorithm2e}
\usepackage{bigints}
\usepackage{eurosym}
\usepackage{graphicx}
\usepackage{booktabs}
\usepackage{lmodern}
\usepackage{microtype}      
\usepackage{nicefrac}       
\usepackage{natbib}

\usepackage{scalerel,stackengine}
\stackMath
\newcommand\reallywidehat[1]{%
\savestack{\tmpbox}{\stretchto{%
  \scaleto{%
    \scalerel*[\widthof{\ensuremath{#1}}]{\kern-.6pt\bigwedge\kern-.6pt}%
    {\rule[-\textheight/2]{1ex}{\textheight}}
  }{\textheight}%
}{0.5ex}}%
\stackon[1pt]{#1}{\tmpbox}
}

\newcommand{\betabf}{\boldsymbol{\beta}}
\newcommand{\thetabf}{\boldsymbol{\theta}}
\newcommand{\lambdabf}{\boldsymbol{\lambda}}
\newcommand{\nbf}{\mathbf{n}}
\newcommand{\vbf}{\mathbf{v}}
\newcommand{\lbf}{\mathbf{l}}
\newcommand{\abf}{\mathbf{a}}
\newcommand{\bbf}{\mathbf{b}}
\newcommand{\qbf}{\mathbf{q}}
\newcommand{\pbf}{\mathbf{p}}
\newcommand{\Rbb}{\mathbb{R}}
\newcommand{\Acal}{\mathcal{A}}
\newcommand{\Ical}{\mathcal{I}}
\newcommand{\Ccal}{\mathcal{C}}
\newcommand{\Scal}{\mathcal{S}}
\newcommand{\betahat}{\reallywidehat{\beta}}
\newcommand{\phat}{\reallywidehat{p}}
\newcommand\given[1][]{\:#1\vert\:}

\DeclareMathOperator{\E}{\mathbb{E}}

\AtBeginEnvironment{quote}{\itshape}

\newenvironment{expert}
{\begin{quote}\itshape\normalsize}
{\end{quote}}

\renewcommand{\headeright}{Preprint}
\renewcommand{\undertitle}{Preprint}

\begin{document}

\title{Assessing Supply Chain Cyber Risks}

\author{Alberto Redondo  \\
        Instituto de Ciencias Matemáticas \\
        Nicolás Cabrera, nº13-15 \\
        28049 Madrid, Spain \\
        \texttt{alberto.redondo@icmat.es} \And 
        Alberto Torres-Barr\'{a}n \\
        Instituto de Ciencias Matemáticas \\
        Nicolás Cabrera, nº13-15 \\
        28049 Madrid, Spain \\
        \texttt{alberto.torres@icmat.es}\And
        David R\'{i}os Insua \\
        Instituto de Ciencias Matemáticas \\
        Nicolás Cabrera, nº13-15 \\
        28049 Madrid, Spain \\
        \texttt{david.rios@icmat.es} \And 
        Jordi Domingo \\
        Blueliv \\
        Aribau 197, 3rd floor\\
        08021 Barcelona, Spain\\
        \texttt{jordi.domingo@blueliv.com}}

\maketitle

\begin{abstract}
Risk assessment is a major challenge for supply chain managers, as it potentially affects
business factors such as service costs, supplier competition and customer expectations. The
increasing interconnectivity between organisations has put into focus methods for 
supply chain cyber risk management. We introduce a general approach to support such activity taking into account various techniques of attacking an organisation and its suppliers,
as well as the impacts of such attacks. Since data is lacking in many respects, we
use structured expert judgment methods to facilitate its implementation. We couple a
family of forecasting models 
to enrich risk monitoring.  The approach may be used to set up risk alarms,
negotiate service level agreements, rank suppliers and identify insurance needs, among other
management possibilities.
\end{abstract}

\section{Introduction}
Earthquakes, economic crises, strikes, terrorist attacks and other events may disrupt supply chain operations with significant impact over the performance of organisations and the availability of their products and/or services. As examples, it
is reported that Ericsson lost 400 million EUR after their supplier's semiconductor plant caught on fire in 2000 \citep{ericssonFire} and that Apple lost many customer orders during a supply shortage of DRAM chips after an earthquake in Taiwan in 2016 \citep{appleChipTaiwan}.

Supply chain risk management (SCRM) has come into place to implement strategies to manage risks in a supply chain with the goal of reducing vulnerabilities and avoid service and product disruptions. As in other risk analysis application areas,
SCRM usually involves four processes: identification, assessment, 
controlling and monitoring of risks \citep{bedford2001mathematical}. \cite{tang2008power} define the field as \textit{the management of risks through coordination or collaboration among supply chain partners to ensure profitability and continuity}. They consider four basic impact mitigation areas: supply, demand, product and
information management. \cite{ritchie2007supply} develop a framework that categorizes risk drivers and integrates risks dimensions in supply chains. \cite{sharland2003impact}, \cite{juttner2005supply} and \cite{zsidisin2008supply} identify key issues in SCRM through surveys, presenting best practices.
\cite{hallikas2004risk} propose a risk management process in network environments giving a more holistic view through a risk matrix approach,
although this type of tool has important shortcomings \citep{anthony2008s}. \cite{thekdi2016supply} introduce dependencies through an input-output model across multiple sectors to assess the social and economic factors associated
with SCRM.
\cite{kern2012supply} illustrate how supply chains would benefit from the capacity of predicting
service unavailability early enough, so as to mitigate interruptions.
\cite{foerstl2010managing}, \cite{bandaly2012supplyghadge} and \cite{ghadge2013systems} develop frameworks that cover identification, assessment, response management and performance outcomes. \cite{curkovic2015managing} identify how companies may manage supply chain risks through FMEA. \cite{heckmann2015critical} determine the core characteristics for supply-chain risk understanding, focusing on the definition of supply chain risk and related concepts such as objective-driven risk, risk exposition, disruptive triggers and risk attitude. \cite{fahimnia2015quantitative} present a review of quantitative methods for SCRM, which are expanding rapidly in the field. \cite{aqlan2016supply} and \cite{fattahi2017responsive} use optimization and simulation to deal with deterministic and stochastic features in SCRM. \cite{qazi2017exploring} capture dependence between risks and risk mitigation strategies using Bayesian Belief Networks. \cite{zahiri2017toward} and \cite{song2017modeling} are recent examples of areas where SCRM has a major impact in real applications. Finally, the role of humans in supply chains is reviewed in \cite{perera2018human}.

Due to the proliferation of cyber attacks
and the increasing interconnectivity  of organisations,
a major issue of recent interest refers to new cyber threats 
affecting supply chain operations
in what we shall call Supply Chain Cyber Risk Management (SCCRM).
As an example, Target suffered in 2013 a major cyber breach
through their air conditioning supplier losing up to 70 million 
credit and debit cards of buyers, 
with massive reputational damage
\citep{target2014}. Another relevant attack was Wannacry which took over,
among many others, Telefonica and the UK NHS producing the unavailability
of numerous services, entailing costs estimated to have reached $\$4$ billion \citep{wanacry2016}. 

We present here a framework for SCCRM. Section \ref{sec:description} presents
a general description, covering models to forecast attacks and their 
impacts and  integrating such information to provide relevant risk indicators. Due to lack of data, we need to rely on expert judgment to assess the involved parameters.
Section \ref{sec:implementation} outlines its implementation. Section \ref{sec:example} covers a numerical example. We end up with some discussion in Section \ref{sec:discussion}.

\section{A Framework for SCCRM based on Expert Judgment}\label{sec:description}
Consider a company $c$ interconnected with its suppliers
$s$, pertaining to a set  $\Scal$.
Both the company and the suppliers are subject to various 
types of attacks $a \in \Acal$. Examples include attacks through botnets or based on stolen login information. Attacks to a supplier could be transferred to 
the company.  
As an example, imagine a case in which one of the company's 
suppliers is infected through malware;
the attacker could then scan the supplier's network and send attacking 
emails to the company, which would be more likely to 
get infected as the received software originates from a legitimate source.

We have access to a threat intelligence system (TIS) \citep{tis2017_2} which
collects data $\nbf^a_{c}$, $\nbf^a_{s}, s \in S$, respectively from the 
company and its suppliers, in connection with various attack vectors $a$. The data could include, e.\,g., the IPs of botnet infected devices or the number of malware infections found. 
The TIS provides also data about the security environment 
(including, for instance, the number of negative mentions in hacktivist blogs), and the
security posture (covering for example the patch cadence or the number of
vulnerabilities).
Based on such data, and other available information, we aim at assessing:
\begin{itemize}
\item the probabilities  that the company's suppliers are attacked;
\item the probability that the company is attacked, either directly or through its suppliers;
\item the impacts that such attacks might induce over the company. 
\end{itemize}
We shall then aggregate such information to facilitate the cyber risk assessment
to the company in relation with its suppliers so as to support 
supply chain cyber risk management decisions.

We describe a general approach to obtain, combine and apply in practice the required
model ingredients. Given the reluctance of organizations to provide data concerning
sufficiently harmful attacks for reputational reasons, we deal with the eventual
lack of data to fit the proposed models by extracting information from cybersecurity
specialists through structured expert judgment techniques
\citep{cooke1991expertsWithout,o2006uncertain}. Here, the term ``sufficiently harmful''
indicates that the attack is relevant in the sense of having caused significant damage to the company
and/or 
its suppliers\footnote{For example, this could correspond to attacks that need to be
declared to a supervisory authority, as in the recent General Data Protection
Regulation, \cite{gdpr}.}.

\subsection{Probability of a Sufficiently Harmful Attack}\label{sec:prob}
We start by describing how to estimate the probability that the supplier, or the company, is successfully
attacked, given the information 
scanned through the TIS.  For the moment, 
we do not include
the security environment and posture in the model for expositional simplicity,
which we cover in Section \ref{sec:env}. We 
undertake the proposed approach for each attack type $a \in \Acal$.

 The attack probability is modeled through $$Pr(y = 1 \given  \betabf, \nbf) = g(f(\betabf, \nbf))$$
where $y = 1$ indicates that the attack was successful; 
$\nbf$ represents the data available, $\betabf$ are parameters
and $f: \Rbb^h \rightarrow \Rbb$  and $g: \Rbb \rightarrow [0, 1]$ are invertible functions, 
where $h$ is the number of severity levels of the corresponding attack vector.
For example, if the security event is related to malware attacks, we could have three severity levels: $n_{1}$ (low level, the malware does not seem potentially harmful);
$n_{2}$ (medium level, the malware could cause damage); and $n_{3}$ (high level, the malware
is potentially very harmful). 
The function $f$ should satisfy the condition $f(\betabf, \nbf) = 0$ if $\nbf = 0$.
Finally, to limit the number of questions to be posed to the experts, we also
require the function $f$ to be 
separable, in the sense $$f(\betabf, \nbf) = \sum_{i=1}^h {f_i( \beta_i, n_i)}.$$ 

 The parameters will be indirectly estimated through expert judgment, for which
 we provide appropriate questions and consistency checks. We use interactive elicitation 
 techniques \citep{clemen1999correlations} to assess from experts, for example, 
 the probability $\phat_i$ given the data $\nbf$ so that 
\begin{equation}\label{kakaejor}
\phat_i = Pr(y=1 \given \beta_i, n_i) = g(f_i(\beta_i, n_i)),
\end{equation}
We then make
\begin{equation}\label{kakaejor2}
\hat{\beta_i } = f^{-1}_i (g^{-1} (\phat_i), n_i)).
\end{equation}
Since $f$ is separable, we only need one question per parameter $\beta_i$.

Among many other possibilities, a choice for the functions
$f$ and $g$ are provided by the logistic regression
model \citep{appliedLogisticRegression2013}, that is,
$f(\betabf, \nbf) = \betabf \cdot \nbf$ and $g(x) = 1/(1 + \exp(-x))$.
It is common to add a bias term $\beta_0$, so the total number of parameters would be
$\, h+1$.
We illustrate next how to obtain the parameters given the attack probabilities for
this specific case in which
\begin{equation}\label{logreg}
Pr(y = 1 \given  \betabf, \nbf) = \frac{1}{1+\exp(-(\beta_{0} + \betabf \cdot \nbf))},
\end{equation}
where $\nbf=(n_{1}, \dots, n_{h})$ is the attack vector and, finally,
$\betabf = (\beta_{1},\dots, \beta_{h})$ are the parameters.

We thus ask the expert to assess the attack probability when $\nbf = 0$,
and, as in (\ref{kakaejor}),
\[
\phat_{0} = Pr(y=1 \given \beta_{0},\betabf, \nbf=[0,\dots,0])=\frac{1}{1+\exp(-\beta_{0})}.
\]
A typical question that could be posed to an expert in this case is:
\begin{expert}
Assume that the TIS has detected no evidence in the network concerning such
infection (i.e., $n_i=0$, $i=1,\dots,h$). What would be the probability $p_{0}$ of actually
suffering a sufficiently harmful attack due to such type of 
infection?
\end{expert}
We then solve for $\beta_0$, as in (\ref{kakaejor2}), to obtain
\[
\betahat_{0}= \log \left( \frac{\phat_0}{1-\phat_0}  \right).
\]
Next, we assess through expert judgment
\[
\phat_{1}=Pr(y=1 \given \beta_{0},\betabf, \nbf=[1,0,\dots,0])=\frac{1}{1+\exp(-(\beta_{0}+\beta_{1}))},
\]
obtaining
$$\betahat_{1} = \log\bigg(\dfrac{\phat_{1}}{1-\phat_{1}}\bigg)-\betahat_{0}.$$
We would extract $\phat_{i}$ from the experts in a similar manner to obtain $\betahat_{i}$, $i=2,\dots,h$.
Finally, we would check for consistency  using assessments such as, e.\,g., $$\phat =
Pr(y=1 \given \beta_{0},\betabf , \nbf=[2,0, \dots,0]),$$ 
and checking whether
\[
\log\bigg(\frac{\phat}{1-\phat}\bigg) \simeq \betahat_{0}+2\betahat_{1}.
\]
If not, we would need to reassess some of the judgments,
modifing the parameters accordingly. 

Besides the attack probabilities, 
for the case of suppliers 
we also need to
assess the probability $q^{a}$ of a type $a$  attack being transferred 
from the supplier to the company.
We define
an attack to a supplier as transferred successfully 
if it is immediately followed by a second attack to the company,
taking advantage from either the information gathered in the
first attack or the compromised infrastructure.
The probabilities of transferring an attack are different
for each of the atta types. Therefore, we elicit them directly from the experts. 
An example of a typical question for assessing such probabilities,
in relation e.\,g., with malware, would be

\begin{expert}
Suppose that there is an attack to a supplier based on 
malware. What would be the probability of the customer suffering
another one, taking advantage of the supplier's attack?
\end{expert}
As before, we introduce consistency checks and interactive
procedures to evaluate
such assessment. We assume that the probabilities are the same for every supplier,
considering that all suppliers are equivalent in that respect, 
thus reducing the cognitive 
load over the experts and avoiding raising the number of questions posed to them.

\subsection{Taking into Account the Security Environment and Posture}\label{sec:env}
We describe now how we incorporate information about the security environment and posture of the supplier and the company within the attack probabilities. Essentially, we introduce indices for the corresponding variables through multicriteria value functions \citep{Gonzalez-Ortega2018} and then apply the approach in Section \ref{sec:prob} to extract the required coefficients.

We define first an index $e$ which assesses the security environment of, say, the supplier
based on the $k$ environment variables captured by the TIS.  Let $e_{i}$ be the $i$-$th$ variable, $i=1,\dots,k$,
rescaled to $[0,1]$. With no loss of generality, assume that the bigger $e_{i}$ is, the worse is the
precieved security environment. We use a multicriteria linear value function, $ \tilde{e} = \sum_{i=1}^{k}\lambda_{i}e_{i}$ with $\sum_{i=1}^{k} \lambda_{i}=1$, $\lambda_{i} \geq 0$, $i=1,\dots ,k $. We determine the $\lambda_{i}$ weights by asking experts to compare
 pairs of security environment contexts leading to a system of equations
\begin{equation}
\begin{aligned}
\delta_{1}^{1}\lambda_{1}&=\delta_{1}^{2}\lambda_{2},\\
&\vdots \\
\delta_{k-1}^{1}\lambda_{k-1}&=\delta_{k-1}^{2}\lambda_{k}.
\end{aligned}
\label{weightsCoefficients}
\end{equation}
For instance, given a reference value $\delta_{1}^{1}$
for the first environment variable, we obtain the first equation 
by asking the expert about the value $\delta_{1}^{2}$
 of the second environment variable such that the following two security environments
 are perceived as equally unsafe by the expert
\[
(\delta_{1}^{1},0,\dots,0) \sim (0,\delta_{1}^{2},\dots,0).
\]
As an example, given that $\delta_{1}^{1}=0.7$,
the expert answer could be 
 $\delta_{1}^{2}=0.3$, leading to the equation $0.7 \lambda_{1}=0.3 \lambda_{2}$.
 As before, we  introduce interactive schemes to obtain the $\delta_{i}$'s and perform consistency checks.

Then, from system \eqref{weightsCoefficients}, we obtain the
$k-1$ equations
\[
\lambda_i = \frac{\delta_{i-1}^1}{\delta_{i-1}^2} \lambda_{i-1} = r_{i-1} \lambda_{i-1}, \quad i=2,\dots,k,
\]
with
\[
r_i = \delta_{i}^1/\delta_i^2,\quad i=1, \dots, k-1.
\]
Taking into account that $\sum_{i=1}^{k}\lambda_{i}=1$, we solve for $\lambda_1$ to obtain
\[
\lambda_1\Big(1 + \sum_{i=1}^{k-1}{\prod_{j=1}^{i}{r_j}}\Big)= 1,
\]
so that
\[
\lambda_1 = \frac{1}{1 + \sum_{i=1}^{k-1}{\prod_{j=1}^{i}{r_j}}},
\]
and
\[
\lambda_i = \prod_{j=1}^{i-1}{r_j} \lambda_1,\quad i=2,\dots,k.
\]

We proceed in a similar manner to aggregate the security posture, defining an index
$l$ and assessing such posture through a linear multicriteria value function
\[
\tilde{l} = \sum_{i} v_{i}l_{i},
\]
with $\sum_{i} v_{i}=1$ and $v_{i} \geq 0\; \forall i$,  where $l_{i}$ is the $i$-$th$ security posture variable, conveniently scaled.

Once we have built the 
environment and posture 
value functions, we assess the 
corresponding $\beta$ parameters.
We adopt common parameters for all attack types. For example,
 for the specific case of a harmful attack with $h$ levels of infection, we just extend the
 attack vector $\widetilde{\nbf} = [\nbf, \tilde{l}, \tilde{e}]$, where $[\abf, \bbf]$ denotes the
 concatenation of vectors  $\abf $ and $\bbf$, and plug it into equation \eqref{kakaejor} or \eqref{logreg}. 

Then, we set a reference value for $\tilde{l}$, say $v_{1}$, which we 
associate with $\lbf =[1,0,\dots,0]$ and ask about the corresponding
probability $\phat_{h+1}$, when, e.\,g.,  $n_i=1$, $i=1, \dots, h$ and
$\tilde{e}=0$. For the specific case with the logistic regression model, equation \eqref{logreg} becomes
$$
\phat_{h+1} = \frac{1}{1 + 
\exp(-(\beta_{0}+\beta_{1}1+\dots+ \beta_{h}1+\beta_{h+1}v_1 +\beta_{h+2} 0))}, 
$$
and, similarly, in the general one.
Since we have already elicited $\betahat_{0}$, $\betahat_{1}$, $\dots$, $\betahat_h$,
Section 2.1, we easily obtain $\betahat_{h+1}$, through
$$\betahat_{h+1}=\frac{1}{v_{1}}\bigg(\log\bigg(\frac{\phat_{h+1}}{1-\phat_{h+1}}\bigg)-\sum_{i=1}^{h}{\betahat_i}\bigg).$$
We proceed similarly to obtain $\beta _{h+2}$, introducing the corresponding consistency checks.

 As mentioned, we implement this approach for all attack types, both for the company and its suppliers.

\subsection{Impacts over the company and its suppliers}\label{sec:impact}
We describe the models used to predict the impacts of attacks over the company. The relevant impacts
might vary across the organisation.
In our supply chain application area we have included: the suppliers' service unavailabilities,
 as they would induce a cost in the company due to lacking such service;
 the company's service unavailability, which would also induce a cost,
 typically, higher than the earlier ones;
 and, finally, the loss of company reputation
 which might induce, in turn, a loss of customers. We use the same distributions for all attack types. \cite{hubbard2016measure} discuss how to measure other impacts 
 potentially relevant in cybersecurity.
\subsubsection{{\it Supplier and Company Unavailability}}\label{sec:downtime}
We focus first on supplier and company service unavailability, given a sufficiently
harmful attack. Their durations
will be designated by $i_s$ and $i_c$, respectively. We model the corresponding downtimes
through  non-negative parametric distributions
\[
f(i_s \given \thetabf_s) \sim P(\thetabf_s); \,\,\,\,  f(i_c \given \thetabf_c) \sim P(\thetabf_c).
\]

Given the lack of data, we aim at obtaining estimates of the parameters through expert judgment.
 For this, we ask the experts for several quantiles $\qbf=(q_{1},\dots,q_{n})$ of the corresponding distributions and estimate the parameters $\thetabf$  by solving
\begin{equation*}
\min_{\thetabf}\;|| \qbf - \text{cdf}_P(\pbf,\, \thetabf)||_{2}^{2} ,
\end{equation*}
where $\text{cdf}_P(\cdot,\,\thetabf)$ designates the cumulative distribution function of the distribution $P$ with parameters $\thetabf$. This leads to the corresponding optimal parameters $\thetabf^*$. As an example, if the distribution had two parameters, we could use two quantiles,
say the first and third quartiles, with $\pbf = [0.25, 0.75]$. As before, we also perform consistency checks based on other quantiles and use interactive procedures to obtain the involved quantiles. 

When required, we may approximate the downtimes through the expected value of the distributions
\[
\bar{i}_s =\E[f(i_s \given \thetabf_s^*)];\,\,\, 
\bar{i}_c =\E[f(i_c \given \thetabf_c^*)].
\]
We undertake this approach for all the suppliers and the company.

\subsubsection{{\it Reputation}}\label{sec:reputation}
We consider now the impact of an attack over the company's reputation.
We assume that reputational impacts occur only if there is a direct
attack to the company.
 There is no natural attribute that allows us to assess
 reputation loss, see the discussion in
 \cite{hubbard2016measure}. Our focus is therefore in its
 business consequence which we consider to be the loss in market share induced by a 
 harmful attack over the organisation.

 Let us designate by $d$ the proportion of customers abandoning
 to a competitor due to the
 incumbent loss of reputation, which we model as a parametric distribution with support 
 in $[0,1]$.
   We proceed in a similar fashion to Section \ref{sec:downtime}, by asking several quantiles 
   to experts and, subsequently, approximating the parameters based on a least squares
   cdf approximation, after appropriate consistency checks. When required, the expected
   proportion of customers lost could be approximated through the expected value.

\subsubsection{{\it Aggregating Impacts}}\label{sec:cost}
We describe now how we aggregate all relevant impacts over
the company. For such purpose,  we would
require $\tau$, the market share for the company; $\eta$, the (monetary) market
size; and $\kappa_{s}$ and $\kappa_{c}$, the cost per hour of supplier and company
service unavailability, respectively. We may assess them from data and/or experts.

First, we compute the downtime cost of the supplier $s$ and company $c$ after a 
sufficiently harmful attack through
%
$$c_{i_s} = \kappa_{s} \times i_{s},$$ 
$$c_{i_c} = \kappa_{c} \times i_{c}.$$
%
On the other hand, the reputational cost after a successful
attack is approximated through the cost associated with clients abandoning
the company, which would be
$$c_d = d \times \tau  \times \eta.$$
Recall now that there are three types of attacks:
\begin{enumerate}
\item Direct attacks to the company. The cost in this case is $c = c_d + c_{i_c}$.
\item Attacks to the supplier that disrupt its service but are not transferred to the company. Their costs are $c= c_{i_s}$.
\item Attacks to the supplier that disrupt its service and are transferred to the company. The cost would be $c =  c_d + c_{i_c} + c_{i_s}$.
\end{enumerate}
To cater for the company's risk attitude, we 
may use a constant risk averse utility model \citep{Gonzalez-Ortega2018},
\[ u(c)=   (1-e^{-\rho c})/\rho ,
         \]
where $\rho$ is extracted from experts within the company, as there explained.

When necessary, we shall use the expected costs 
\[ \bar{c}_{i_s} = \kappa_{s} \times \bar{i}_{s}, \;\; 
\bar{c}_{i_c} = \kappa_{c} \times \bar{i}_{c}, \;\;
\bar {c}_d = \bar {d} \times \tau \times \eta .
\]

\subsection{Model Outputs}
We derive now the model outputs that may be relevant for risk management, including attack probabilities, expected impacts and expected utilities.
\subsubsection{{\it Attack Probabilities}}
For the incumbent company and its suppliers, recall that the probability of a type $a \in \Acal$ security event resulting in a sufficiently harmful
attack is
\begin{equation}
p_i^a = Pr(y = 1 \given  \betabf^a, \nbf^a_i) = g(f(\betabf^a, \nbf_i^a)),
\end{equation}
where $\nbf^a_i$ represents the count for the attack vector $a$, including the environment and posture indicators, and $i \in \{c,\,s\}$ ($c$ is the company and $s$ a supplier).
Assuming that the attack types are independent, the probability of a direct attack to the company will be approximated through
\begin{equation*}\label{AP}
\text{AP}_c = \sum_{k = 1}^{K_\Acal}\; \sum_{\Ical \in \Ccal_{\Acal, k}} \left( \prod_{a \in \Ical}{p_c^a}  \prod_{a \in \Acal \setminus \Ical} {(1 - p_c^a)}  \right),
\end{equation*}
where $\Ccal_{\Acal, k}$ is the set of all possible combinations of $k$ elements taken
from $\Acal$ and $K_\Acal \leq |\Acal|$ is the maximum number of simultaneous attacks 
taken into account. A typical value for this parameter would be 3, as very rarely organisations 
are simultaneously attacked through more than three vectors \citep{reputation_risk}.

For each supplier $s \in \Scal$, we approximate the induced attack probability 
from $s$ to $c$ through 
\begin{equation*}
  \text{IAP}_c^s = \sum_{k = 1}^{K_\Acal}\; \sum_{\Ical \in \Ccal_{\Acal, k}} \Bigg[ \underbrace{\Bigg( \prod_{a \in \Ical}{p_s^a}  \prod_{a \in \Acal \setminus \Ical} {(1 - p_s^a)}\Bigg)}_{\text{Probability of direct attack}}
  \underbrace{\Bigg(1 - \Big( 1 - \prod_{a \in \Ical\vphantom{\setminus}}{(1 - q^a)} \Big)\Bigg)}_{\text{Probability of transferred attack}} \Bigg],
\end{equation*}
recalling that $q^a$ is the probability of an attack of type $a$ being transferred from $s$ to $c$.

Finally, we approximate the global attack probability to the company through
\begin{equation}
  \text{GAP}_c \simeq \text{AP}_c + (1 - \text{AP}_c) 
   \sum_{k = 1}^{K_\Scal}\; \sum_{\Ical \in \Ccal_{\Scal, k}} \left( \prod_{s \in \Ical}{\text{IAP}_c^s}  \prod_{s \in \Scal \setminus \Ical} {(1 - \text{IAP}_c^s)}  \right),
\label{GAP}
\end{equation}
where $\Scal$ is the set of all suppliers of the company $c$ and $K_\Scal \leq |\Scal|$
is the maximum number of suppliers that can reasonably transfer an attack over the
same time period. In (\ref{GAP}), we observe that when an attack
is direct to the company it supersedes the indirect attacks as its consequences will 
be typically much more important.

\subsubsection{{\it Risk Measures: Expected Impacts of Attacks}}
\label{subsubsec:riskMeasuresExpectedImpacts}
Recall that we are assuming that if an attack is successfully transferred from a supplier, there are unavailability and reputational costs, whereas if the attack is not transferred, 
we only consider supplier unavailability costs. We now approximate the expected impacts. 

The first one refers to the expected impact due to direct attacks to the
company, and is expressed as
\begin{equation*}\label{risk}
\text{R}_c = AP_{c}(\bar{c}_{d} \times \bar{c}_{i_s}).
\end{equation*}
The impact induced through attacks to the supplier $s$ is specified as  
\begin{multline*}
\text{IR}_c^s = \sum_{k = 1}^{K_\Acal}\; \sum_{\Ical \in \Ccal_{\Acal, k}} \Bigg\{ \Bigg( \prod_{a \in \Ical}{p_s^a}  \prod_{a \in \Acal \setminus \Ical} {(1 - p_s^a)}\Bigg)
  \Bigg. \Bigg. \Bigg[\overbrace{\Bigg(1 - \prod_{a \in \Ical}{(1 - q^a)} \Bigg)(\bar{c}_{i_s})}^{\text{Attack not transferred}} +
\\
\Bigg. \Bigg. \underbrace{\Bigg(1 - \Bigg(1 - \prod_{a \in \Ical}{(1 - q^a)} \Bigg)\Bigg)(\bar{c}_d + \bar{c}_{i_c} + \bar{c}_{i_s})}_{\text{Attack transferred}}  \Bigg] \Bigg\}.
\label{ind_risk}
\end{multline*}
Finally, the total impact would be
\begin{equation*}\label{total_risk}
\text{TR}_c = R_c + \sum_{s \in \Scal}{\text{IR}_c^s}.
\end{equation*}

In a similar fashion, we may consider approximations to expected utilities replacing the previous impacts by the corresponding utilities.

\subsection{Forecasting Risk Indicators}\label{sec:dlm}
The previous approach is used periodically over time based on
collecting data through the TIS and aggregating the results to assess supply
chain cyber risks. As a relevant complement, observe that the proposed approach
focuses on studying several
risk indicators $X_j$ (attack probabilities, expected impacts, expected utilities)
to monitor SCCR at the company, in reference to time $j$.
The ensuing analysis focuses on just one of the indicators, but applies
to all of them. $D_j$ represents the data available until time
$j$. $X_{j+1} | D_j$ represents a forecasting model for the risk index at
time $j+1$ and summarises all information available at time period $j$ concerning such index.

We employ Dynamic Linear Models (DLMs) to support forecasting tasks in risk monitoring.
We briefly sketch the basic DLM results we use. For further details, see \cite{west2013bayesian} and \cite{dynamicLinearModelsPetris2009}.
 We adopt the general, normal DLM with univariate observations $X_{j}$, characterised
 by the quadruple $\{F_{j},G_{j},V_{j},W_{j}\}$, where, for each $j$, $F_{j}$ is a known
 vector of dimension $m\times1$, $G_{j}$ is a known $m\times m$ matrix, $V_{j}$ is a known
 variance, and $W_{j}$ is a known $m\times m$ variance matrix.
 The model is written as
\begin{align*}
\theta_{0}|D_{0}          &\sim N (m_{0}, C_{0}), \\
\theta_{j}|\theta_{j-1}   &\sim N (G_{j}\theta_{j-1}, W_{j}), \\
X_{j}|\theta_{j}          &\sim N (F_{j}^{\prime}\theta_{j}, V_{j}).
\end{align*}

Because of the relative stability of the type of series considered, 
for modelling purposes we use a trend (second order polynomial) DLM which is a constant $F_j$ and $G_j$ specification with
\begin{equation*}
F = \left[\begin{array}{cc}
1&0
\end{array} \right]
\quad
\text{and}
\quad
G = \left[ \begin{array}{cc}
1&1\\
0&1\\
\end{array} \right].
\end{equation*}
\cite{west2013bayesian} summarise the basic features of DLMs for forecasting purposes that we use. They are based on the one-step ahead predictive distributions which, for each $j$, have normal distribution
\[
X_{j}|D_{j-1}\sim N(f_{j}, Q_{j}),
\]
with mean $f_{j}$ and variance $Q_{j}$ recursively defined. $k$-steps ahead forecasts are
also based on normal models and will also be used below.

\subsection{Uses}
We sketch the main uses of the above outputs which facilitate the
implementation of supply chain cyber risk management strategies.

\paragraph{Risk monitoring}
As \cite{kern2012supply} illustrate, predicting supply chain disruptions well in 
advance greatly benefit operations.
Indeed, a major activity in operational risk management refers to monitoring risk levels and
advice sufficiently in advance on when potentially dangerous situations might arise.

We base risk monitoring on the above mentioned 
forecasting models. For this, we use the point forecasts
at time $j$ for the risk indicator
given by $\E[X_j|D_{j-1}] = f_j$ and the interval forecasts determined by $[l_j = f_j - z_{1-\alpha/2}Q_j^{1/2},\, u_j = f_j + z_{1-\alpha/2}Q_j^{1/2}]$,
where $u_j$ and $l_j$ respectively represent the upper and lower bounds of the interval;
 $z_{1-\alpha/2}$ is the $1-\alpha/2$ quantile of the standard normal distribution; and, finally, $\alpha$ is the desired probability level of the predictive interval.
 We then issue an alarm about an unexpected
 increase, or decrease, in the incumbent risk
 indicator when the corresponding next observation $x_{j}$
 does not fall in the predictive interval $[l_j,\, u_j]$. If $x_{j} > u_{j}$, the risk seems much worse than expected possibly flagging an issue; if $x_{j} < l_{j}$, the risk is better than expected suggesting possibly improved security practices. 
  Should this happen repeatedly over time, the alarm could be modulated to 
  accordingly increase awareness.

   Another type of monitoring alarms
   may be raised when the predictive interval captures a sufficiently high risk level $y$. For this, we perform
   predictions $k$ steps ahead based on standard DLM forecasting results to try to forecast sufficiently in advance
   critical risk issues. For example, we may try to detect whether $Pr (X_{j+k} \geq y) $    will be 
   sufficiently high for a certain $k$ and, if so, raise a warning.
\paragraph{Supplier relations management} 
Another important activity in SCRM refers to managing relations with suppliers. 
A key issue relates with ranking suppliers. For this purpose,
we may consider the induced risks, the induced expected impacts
or the induced attack probabilities of various suppliers over the company.
For example, we could say that supplier $s_1$ is preferred to supplier $s_2$ 
if its induced expected impact is smaller, that is, if
 $ \text{IR}_c^{s_1} \leq \text{IR}_c^{s_2}$. This supports identifying and dealing
 with riskier suppliers for communication purposes,
 increasing transparency and forecasting critical situations. As a result of this process
 we may choose between the less risky of two suppliers which provide the same service or product with similar cost.
 This also serves to setup benchmarks between different groups of suppliers.

Another important supplier relations management activity refers to
negotiations of service level agreements (SLA) between the company and a supplier of its choice. Using the same indices as before, SLAs
could be based on the company requiring the supplier to preserve its induce
risk indicator $X_j$ to
remain below a certain maximum level $m_j$, as well as requiring the 
supplier, possibly through a third party, to monitor and forecast whether
such maximum acceptable level has been or will be attained. 
For example, the company could require the supplier to
preserve $\text{IR}_c^{s} \leq m$, over time. 
Repeated violations of the agreed maximum risk 
level could lead to a contract breach and a penalty, thus
incentivising a supplier into
better managing cybersecurity.
\paragraph{Cyber insurance}
The above risk measures allow us to properly apportion the cyber risks to
which a company is subject to, including those related with third parties, thus facilitating the
negotiation of insurance contracts, both with the suppliers and an insurer, as well 
as introducing new dynamic insurance contracts. For example, the
company could apply for an insurance premium reduction if $\text{TR}_c$ is preserved below a
certain agreed level throughout a year; or we could define a, say, weekly premium, based on
the dynamic assessment of the level of various risk indicators. The SLA 
mechanisms described earlier may be seen also
as an insurance mechanism between the company and its suppliers.
\section{Implementation}\label{sec:implementation}
We describe how we implement the proposed approach and couple it with a commercially available TIS.
Essentially, we first calibrate several experts concerning their cybersecurity knowledge. Then
we obtain their judgments and combine them to obtain the corresponding parameters. Finally, during operation, we assess attack probabilities and risks, issue alarms if required and 
forecast risks for the next periods. This entails some data preprocessing.
\subsection{Security Expert Calibration}
We obtain judgments from $m$ experts which we calibrate using
Cooke's (1991) classical model. We first ask the experts several general questions
related with cyber security. As an example, one of the questions we used is:
\begin{expert}
Which was the number of new ransomware types over the last year?
\end{expert}
The experts' answers describe the intervals that, they believe, cover with high probability the value, as well as their median. Based on such assessments, we obtain the expert scores $(\omega_{i})_{i=1}^{m}$, $\omega_{i}\geq 0$, $\sum_{i=1}^{m} \omega_{i}=1$.

\subsection{Security Expert Assessment. Generic Questions}
Once we have calibrated the experts, we extract the
relevant attack probabilities through a questionnaire, based on hypothetical scenarios in which
sufficiently harmful attacks may occur. For example, one of the questions referring
to a scenario pertaining to a botnet based type of attack was:
\begin{expert}
According to you, what would be the probability of actually suffering a sufficiently
harmful attack through botnet infected devices if the TIS did not detect any of them
in the company's network?
\end{expert}
The $i$-th expert provides a probability $p_{i}$ for the corresponding 
question, which we aggregate through
$$p=\sum\limits_{i=1}^{m} \omega_{i} p_{i},$$
with the weights defined above. Based on this type of assessments we obtain the $\beta$
parameters, covering the stages 
described in Sections \ref{sec:prob} and \ref{sec:env}.
\subsection{Company Expert Assessment. Specific Questions}
We then extract the specific parameters which refer to impacts to the company 
(direct or through the supplier) with a questionnaire.
 These parameters include, among others, the current market share,
 the market size, or the expected number of customers lost after a successful 
 attack.
 Some of them will be based on available data; for others, we could ask
 experts as described above.
 For instance, a question we used is:
\begin{expert}
According to you, what would be the average cost of one hour of service unavailability of such supplier?
\end{expert}
In such a way, we cover the issues described in Section \ref{sec:impact}.
\subsection{Data Preprocessing}\label{sec:exp_smooth}
The data received through the TIS are preprocessed via exponential smoothing, \cite{brown2004smoothing}.
This allows us to control the growth rate of various security indicators and partly mitigate fluctuations associated with random variations. Given $\{x_{j}\}$, the raw data sequence, $r_{j}$ will represent the security indicator at time $j$, and $h$ the smoothing factor through
\begin{equation*}
r_{0}=x_{0}, \quad r_{j}=h \times x_{j} + (1-h)r_{j-1}.
\end{equation*}
Observe that
\begin{equation*}
r_{j} = h \times x_{j} + (1-h)h x_{j-1} + \dots +
(1-h)^{k-2} h x_{j-(k-1)} + (1-h)^{k} h r_{j-k}.
\end{equation*}
Since $(1-h)^{k^{*}}$ will be small enough for
$k^{*}$ sufficiently large, we have
\[
r_{j} \simeq \sum_{i=0}^{k^{*}} h (1- h)^{i} x_{j-i},
\]
which effectively entails preserving the last $k^{*}$ scans $(x_{j},x_{j-1},\dots,x_{j-k^{*}})$
and consolidating them in $r_{j}$.

\subsection{Operation}
With the above information, we are capable of putting the proposed approach under 
operation. We summarise first all the 
required information, after having calibrated the experts:
\begin{enumerate}
\item The coefficients $\lambdabf$, $\vbf$, and $\betabf$ are obtained (indirectly) from the experts as in Sections 
2.1, 2.2 and 3.2. 
\item The probability $q^a$ of each attack type transferring from the supplier to the customer is obtained directly from the experts as described in Section 2.1.
\item The information needed to compute the downtime costs of the company and supplier, $c_c$ and $c_s$, must be provided by the company. This involves assessing every supplier $s$ to obtain estimates of its downtime distribution $i_s$, as well as $i_c$.
\item The information needed to compute the reputational cost $c_d$ is obtained from
experts within the company.
\end{enumerate}

Once we have those parameters, we may start operations by essentially scanning through the TIS, performing the risk computations, updating the forecasting models and issuing alarms if required.
The suppliers and company data obtained periodically by scanning the network through the TIS are preprocessed, Section \ref{sec:exp_smooth}. In summary, at every time step $j$ we perform the following computations:
  \begin{enumerate}
      \item Scan the network through the TIS in search for attack, posture and environment vectors $\nbf$, $e$ and $l$, for the company and its suppliers.
      \item Preprocess the data (exponential smoothing and scaling).
      \item Assess the attack probabilities $\text{AP}_c$, $\text{IAP}_c^s$ and $\text{GAP}_c$.
      \item Assess the risks $\text{R}_c$, $\text{IR}_c^s$ and $\text{TR}_c$.
      \item Issue alarms, if required.
      \item Display risk outputs.
      \item Compute risk forecasts for next periods, as required.
  \end{enumerate}
  We sketch next a numerical example using simulated data.
\section{A Numerical Example}\label{sec:example}
Assume that we scan information regarding four attack vectors, $|\Acal| = 4$. For the first security vector, one expert provided attack probabilities $\phat_{0} = 0.05$, when the scan did
not detect any infected devices, and $\phat_{1} = 0.25$, when the scan detected
$1\%$ of infected devices in the company's network. Then, using the basic logistic regression model, we obtain
\[
\betahat_{0} = \log\bigg(\frac{\phat_{0}}{1-\phat_{0}}\bigg) = -2.94,
\]
and
\[
\betahat_{1} = \log\bigg(\dfrac{\phat_{1}}{1-\phat_{1}}\bigg)-\betahat_{0} = 1.85.
\]
If we detect now that $0.075\%$ of the devices are actually infected at $t=1$, we estimate 
the attack probability of that specific security event at
$$p_c = \frac{1}{1+\exp(2.94 - 1.85 \times 0.075)} = 0.057.$$

We show in Table \ref{tab:prob}, columns 1-3, the probabilities of each of the security events
for the company and two of its suppliers $s_1$ and $s_2$. These probabilities are obtained
through equation \eqref{logreg}, as demonstrated above for the company and the first
security event. We also show in the fourth column the probability of an attack
being transferred from a supplier to the company for the four vectors.

\begin{table}[htbp]
\centering
\begin{tabular}{lrrrr}
\toprule
 &   $p_c$ &  $p_{s_1}$ &  $p_{s_2}$ & $q$ \\
\midrule
atk0 & 0.057 & 0.468 & 0.383 & 0.103 \\
atk1 & 0.187 & 0.164 & 0.350 & 0.107 \\
atk2 & 0.131 & 0.166 & 0.143 & 0.056 \\
atk3 & 0.236 & 0.481 & 0.200 & 0.084 \\
\bottomrule
\end{tabular}
\caption{Probabilities of successful attack for four attack types (company, supplier 1, supplier 2, transfer)}\label{tab:prob}
\end{table}

Given the above, we compute the direct attack probability to the company over
the next period, $\text{AP} = 0.491$; the attack probabilities induced by the suppliers
$\text{IAP}_1 = 0.111$, $\text{IAP}_2 = 0.098$; and, finally, the global attack
probability, $\text{GAP} = 0.592$. Clearly having two suppliers increases the chances for
the company of receiving attacks. Should both suppliers offer the same service,
say, both were Internet providers, we could take into account their IAPs to
decide whether to contract the services from one or the other.
In the example, we might be more inclined to work with the second
supplier as it seems less likely to receive an attack through it.

We transform the previous probabilities into expected costs, assuming the
following information:
the average cost per hour of downtime of our company is $20$k EUR; the market size 
is $2922$ billion EUR;  and, finally,
the current market share of the company is $18$\%. In addition, 
the company estimates that one hour of service unavailability of both suppliers costs
$25$k EUR. The first and third quartiles for the downtime distributions are
$2$ and $6$ hours for the company, and $1$ and $4$ hours for the suppliers.
Finally, the first and third quartiles for the distribution of the proportion of lost customers after a
successful attack are assessed as $0.01$ and $0.05$

\begin{figure*}[hbtp]
\centering
\includegraphics[width=0.9\textwidth]{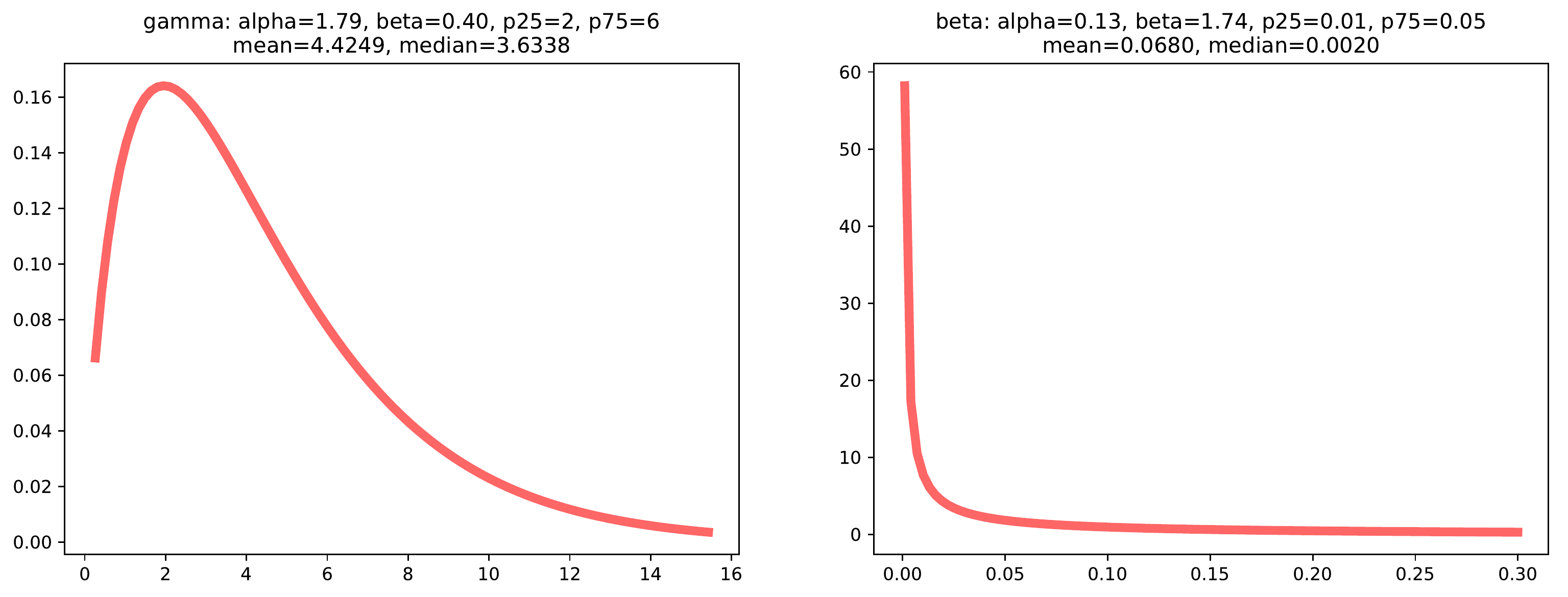}
\caption{Distributions of company downtime duration (left) and
proportion of  customers lost (right)}\label{fig:dist}
\end{figure*}

With the previous information, using the procedure in Section
\ref{sec:downtime},  we estimate 
the distributions of downtime durations. We use Gamma distributions.
As a consequence, the downtimes are modeled as a Gamma$(1.79,\, 0.40)$ for the company and Gamma$(1.21,\,0.42)$ for the suppliers. 
Similarly, using the procedure in Section \ref{sec:reputation}
we estimate the proportion of lost customers based on a Beta distribution,
 leading to a Beta$(0.13,\, 1.74)$, shown in Figure \ref{fig:dist}. 

We aggregate the previous costs and estimate the expected direct cost for
the company as $\text{R} = 517.16$k EUR; the expected cost induced by the
suppliers as $\text{IR}_1=116.73$k, $\text{IR}_2=103.38$k EUR; and, finally, the total
expected cost as $\text{TR} = 737.27$k EUR. Note, again, that the first supplier
seems worse as its entailed expected loss is bigger. 
\begin{figure}[htbp]
\centering
\includegraphics[scale=0.54]{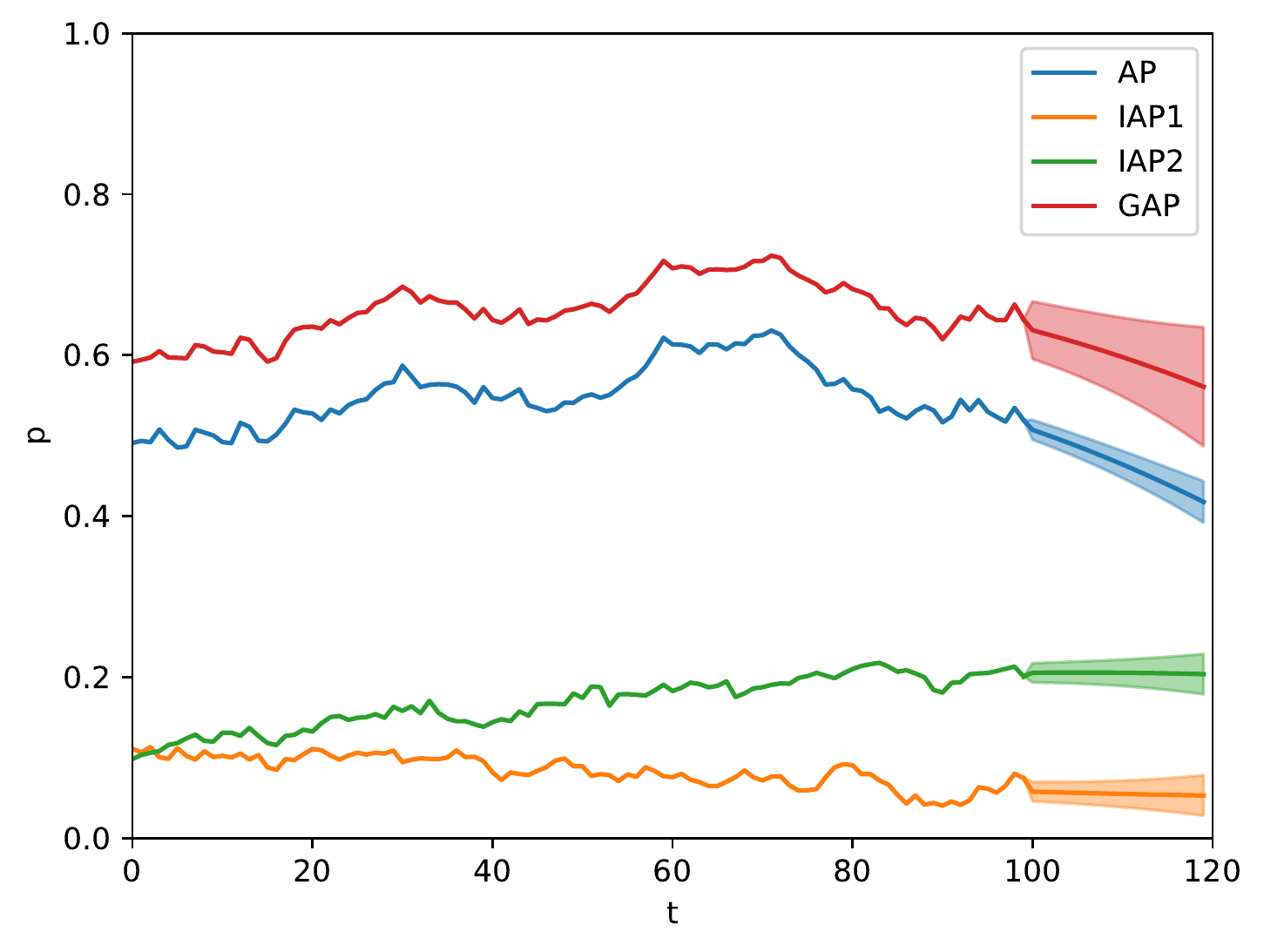}
\caption{Evolution of AP, IAP and GAP}\label{fig:AP}
\end{figure}

The procedure described above would be run periodically, obtaining updated 
values
for the probabilities and costs. An example for $T=100$ time steps is shown in Figure
\ref{fig:AP}, where we plot the evolution of the attack probability, the induced
attack probabilities and the global attack probability. We can appreciate how,
at $t=0$, we would prefer supplier 2 over 1, since it seems to induce
less risk over the company. However, the probability induced by supplier 2 gets worse over time, reverting the situation. We also fit DLMs
for the four probability indices, as described in Section \ref{sec:dlm}. This allows us to forecast the different attack
probabilities $k$-steps ahead. Figure \ref{fig:AP} shows the expected value of the predictive distribution $X_{100+k}|D_{100}$ for $k=1,...,20$, and the corresponding $95\%$ predictive intervals.
 
\section{Discussion}
\label{sec:discussion}
As shown by the  recent emergence of several products in the market,
SCCRM is a very
relevant and 
current managerial problem given the increasing 
importance of cyber attacks and the interconnectivity of organisations in modern economy
and society.
This motivated our approach to support SCCRM.
Given the reluctance of companies to release attack data,
 we have described how we may  extract knowledge from security experts to obtain
 the parameters required to assess risk scores, based on information coming from a
 TIS in relation with
  attack vectors as well as the security posture and environment of a company and its suppliers.
   Besides, we have incorporated forecasting models that allow us to monitor risk predictively and issue alarms. We may also use the provided information to rank suppliers,
   negotiate SLAs or use them for insurance purposes.
   The approach has been implemented in Python routines and integrated with a TIS. 

There are several ways to further advance this work. For example, should data about attacks be revealed, we could introduce schemes to learn about the involved parameters using our assessments as priors, through Markov Chain Monte Carlo procedures, see e.\,g. French and Rios Insua (2000). We could also incorporate additional impacts such as loss of productivity, loss of revenue or the increase of working hours. The model might also consider collaboration among suppliers to mitigate consequences when a service is disrupted or monitoring information sharing between suppliers to avoid unfair competition. Finally, we have included only direct suppliers to the company but we could also consider suppliers of suppliers, and beyond.

\subsubsection*{Acknowledgements}
This work has received funding from the European Union’s H2020 Program for
Research, Technological Development and Demonstration under grant agreement $740920$; the Spanish Ministry of Economy and Innovation
programme MTM2017-86875-C3-1-R; the European Cooperation in Science and Technology (ESF-COST); and the AXA-ICMAT Chair on Adversarial Risk Analysis.

\bibliographystyle{spbasic}
\bibliography{references}

\end{document}